\documentclass[runningheads]{llncs}
\usepackage[T1]{fontenc}
\usepackage{lmodern}

\usepackage[utf8]{inputenc} %
\usepackage[T1]{fontenc}    %
\usepackage{hyperref}       %
\usepackage{url}            %
\usepackage{booktabs}       %
\usepackage{amsfonts}       %
\usepackage{nicefrac}       %
\usepackage{microtype}      %
\usepackage{lipsum}         %
\usepackage{xcolor}         %
\usepackage{tabularx,colortbl}
\usepackage{graphicx}       %

\usepackage{arydshln}       %
\usepackage{rotating}       %
\usepackage{array}
\usepackage{multirow}
\usepackage{graphicx}
\usepackage{float}
\usepackage{nicematrix}
\usepackage{todonotes}
\usepackage{lscape}
\usepackage{times}
\usepackage{ifthen}
\newboolean{combinedversion} 

\usepackage{xr}
\makeatletter
\newcommand*{\addFileDependency}[1]{%
  \typeout{(#1)}
  \@addtofilelist{#1}
  \IfFileExists{#1}{}{\typeout{No file #1.}}
}
\makeatother

\newcommand*{\myexternaldocument}[1]{%
    \externaldocument{#1}%
    \addFileDependency{#1.tex}%
    \addFileDependency{#1.aux}%
}

\setboolean{combinedversion}{true}   
\ifthenelse{\boolean{combinedversion}}{
\newcommand*{\appref}[1]{\ref{#1}}
}{
\myexternaldocument{theappendix}
\newcommand*{\appref}[1]{\ref*{#1}}
}

\title{CAD Models to Real-World Images: A Practical Approach to Unsupervised Domain Adaptation in Industrial Object Classification}
\titlerunning{Practical Approach to Unsupervised Domain Adaptation}

\newcommand\myparagraph[1]{\parindent0pt\textbf{#1}\;}

\author{
  Dennis Ritter
  \inst{1} \and
  Mike Hemberger
  \inst{2} \and
  Marc Hönig
  \inst{3} \and  
  Volker Stopp
  \inst{3} \and
  Erik Rodner
  \inst{4} \and
  Kristian Hildebrand
  \inst{1}
}

\institute{Berliner Hochschule für Technik \and
nyris GmbH \and topex GmbH \and
KI-Werkstatt/FB2, University of Applied Sciences Berlin
}

\begin{document}
\maketitle

\begin{abstract}
    In this paper, we systematically analyze unsupervised domain adaptation pipelines for object classification in a challenging industrial setting.
    In contrast to standard natural object benchmarks existing in the field, our results highlight the most important design choices when only category-labeled CAD models are available but classification needs to be done with real-world images.
    Our domain adaptation pipeline achieves SoTA performance on the VisDA benchmark, but more importantly, drastically improves recognition performance on our new open industrial dataset comprised of 102 mechanical parts. We conclude with a set of guidelines that are relevant for practitioners needing to apply state-of-the-art unsupervised domain adaptation in practice. Our code is available at \url{https://github.com/dritter-bht/synthnet-transfer-learning}.

\end{abstract}

\section{Introduction}
\begin{figure}[tb]
    \includegraphics[width=\linewidth]{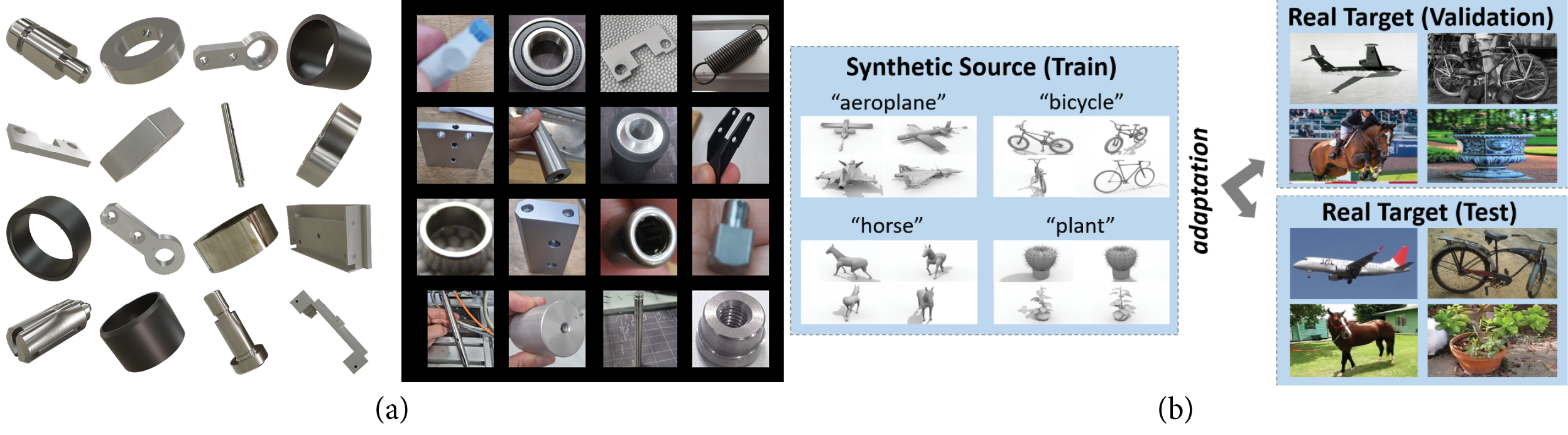}
    \centering
    \caption{
    (a): Our Topex-Printer dataset contains rendered and real images from 102 machine parts (Sect.~\ref{sec:topex_printer_dataset}). (b): The VisDa-2017 challenge tests UDA model performance under simulation-to-real domain shifts~\cite{visda2017}.}
    \label{fig:topex-visda-example}
\end{figure}

Recognizing machine parts requires in-depth industrial domain knowledge. However, particularly in engineering, machine-specific specialists are often needed to identify components without prolonged research, making it challenging for customers of machine manufacturers to independently identify the parts of their machines.
Automatic visual recognition seems therefore a straightforward solution to apply. However, complex machines typically comprise hundreds or even thousands of individual parts. Generating and labeling sufficient images of each component for training is often too costly. In contrast, companies own the computer-aided design (CAD) data of the parts, which can be rendered with any parameters and in any quantity. Consequently, our goal (Fig. \ref{fig:topex-visda-example}) is to use CAD data and train a classifier with adaptation techniques from rendered 3D objects (source domain) that can be applied to real-world images (target domain).

Our proposed contribution is twofold: First, we present a comprehensive guide designed to facilitate future research in surpassing SoTA performance (MIC~\cite{mic2022}) on the VisDA classification challenge benchmark. We analyze the performance enhancements and their impact at the different stages of our domain adaptation (DA) pipeline, providing a blueprint from a wide range of methods already present in the vast existing literature (Sect.~\ref{sec:eval}). Second, we introduce a new open dataset characterized by minimal inter-class distances, offering a novel challenge for unsupervised domain adaptation (UDA) research (Sect.~\ref{sec:topex_printer_dataset}).

Specifically, we use publicly available models pretrained on the ImageNet22K (IN22K) dataset~\cite{in2009} and continue with linear probing using only source domain data to tune the classification head as initialization for further training (similar to \cite{lpft2022}).
We continue training in an unsupervised domain adaption (UDA) setting, i.e. no labels for target domain data available, applying CDAN~\cite{cdan2017} and MCC~\cite{mcc2019}.
We test our approach with the VisDA-2017 image classification challenge dataset \cite{visda2017} and our self-made \emph{Topex-Printer} dataset (Sect.~\ref{sec:topex_printer_dataset}) shown in \figurename~\ref{fig:topex-visda-example}.

\section{Related Work}
\label{sec:rewo}

Adversarial training, which encourages domain-invariant image features, is a key approach in image-based DA techniques. Originally introduced in~\cite{dann2015}, it adapts the GAN concepts of \cite{GAN2014} for DA tasks. ADDA~\cite{adda2017} consolidates several approaches into a framework based on adversarial learning.
CyCADA~\cite{cycada2017} applies CycleGAN's~\cite{cyclegan2017} cycle consistency for DA on image classification and semantic segmentation. 
CDAN~\cite{cdan2017} adds a conditional domain discriminator utilizing classifier predictions to assist the DA process. Lastly, SDAT~\cite{sdat2022} uses a \emph{smooth task loss} to stabilize adversarial training, leading to improved generalization on the target domain.

Beyond adversarial training, discrepancy minimization methods aim to align feature representations, reducing distribution discrepancy between source and target domains. Deep Adaptation Network (DAN)~\cite{dan2015} and JAN~\cite{jan2016} use maximum mean discrepancy (MMD) and joint MMD for feature transfer. 
Contrastive Adaptation Network (CAN)~\cite{can2019} introduces the \emph{Contrastive Domain Discrepancy} (CDD) metric for class-aware alignment. 
Sliced Wasserstein Discrepancy metric (SWD)~\cite{swd2019} is based on the Wasserstein Distance. 
The \emph{Minimum Class Confusion} (MCC) loss~\cite{mcc2019} reduces target domain cross-class confusion. Recently, Masked Image Consistency (MIC)~\cite{mic2022} enforces prediction consistency between masked target images and complete-image pseudo-labels.
Kumar et al.~\cite{lpft2022} suggest an optimized transfer learning scheme that initially updates the classification head, then fine-tunes all parameters—proves to be particularly effective for large distribution shifts in out-of-distribution datasets by preserving pretrained features. Our work adopts this approach, combining CDAN~\cite{cdan2017} and MCC~\cite{mcc2019} for UDA.
While many methods rely on CNNs, recent studies~\cite{tvt2021,pretrainstudy2022} show that Vision Transformer (ViT)~\cite{vit2020} models surpass these. In addition, the benchmark ranking for CNNs does not extend to Transformer models, although pretraining significantly improves domain transfer~\cite{pretrainstudy2022}. For a comprehensive survey of transfer learning, encompassing pretraining and adaptation techniques, refer to \cite{transferability2022}.

We utilize the VisDA-2017 image classification dataset, comprising three subsets: a training set of 150k rendered 2D images from 1,907 3D models, a validation set of 174k real photos from MS COCO~\cite{mscoco2014}, and a test set of 72k real images from Youtube-boundingboxes~\cite{ytbb2017}. Each image is categorized into one of twelve classes. However, as shown later, performance on this dataset already saturates and therefore a novel benchmark is required.

\section{A New Domain Adaptation Benchmark: Topex-Printer}
\label{sec:topex_printer_dataset}

We introduce a challenging dataset for identifying machine parts from real photos, featuring images of 102 parts from a labeling machine.
This dataset was developed with the complexity of real-world scenarios in mind and highlights the complexity of distinguishing between closely related classes, providing an opportunity to improve domain adaption methods.
The dataset includes 3,264 CAD-rendered images (32 per part) and 6,146 real images (6 to 137 per part) for UDA and testing. Rendered images were produced using a Blender-based pipeline with environment maps,  lights, and virtual cameras arranged to ensure varied mesh orientations. We also use material metadata and apply one of 21 texture materials to the objects. We render all images at $512^2$ pixels. Some examples of our rendered images can be seen on the left side of \figurename~\ref{fig:topex-visda-example} (a). The real photo set consists of raw images captured under varying conditions using different cameras, including varied lighting, backgrounds, and environmental factors. More examples are available in the supplementary material. The dataset is publicly available at \url{https://huggingface.co/datasets/ritterdennis/topex-printer/resolve/main/topex-printer.zip}.

\section{Our adaptation pipeline}

We reviewed existing research, analyzing two prevalent stages of DA training. This led to our empirically-backed approach that yielded robust results on the Topex-Printer and VisDA datasets, achieving 93.47\% accuracy on the target domain for the latter, which exceeds the accuracy reported in~\cite{mic2022}. The steps comprise the following:
\begin{enumerate}
\item \textbf{Adapting pretrained models to rendered images:} 
\begin{enumerate}
    \item We start from pretrained models and train a new classification head with source domain data (see~\cite{lpft2022,pretrainstudy2022}). For this, we freeze layers, exchange the class head to the necessary number of classes and tune the class head with source data only (CH).
    \item We executed a fine-tuning across all layers and a hyperparameter search (optimizer, scheduler, learning rate, augmentations) for our DA experiments on source domain data only (FT). 
\end{enumerate}
\item \textbf{Adapting to real-world images with UDA:}
\begin{enumerate}
    \item We use the best parameters from experiments training only with source domain data for our UDA experiments and start training from the checkpoint with the tuned classification head.
    \item We conduct studies on our two datasets with the methods CDAN, MCC, and CDAN-MCC combined and analyze the effect of all our parameters in Sect.~\ref{sec:eval}. 
\end{enumerate}
\end{enumerate}
While these are standard procedures in DA, we lay out the most important aspects for the single steps in the next sections.

\subsection{Adapting pretrained models to rendered images}
\label{sec:dg_experiments}
We conduct transfer learning on various models (ViT~\cite{vit2020}, Swinv2~\cite{swinv22021} and DeiT~\cite{deit2021}, please refer to the supplementary material for version details), pretrained on IN22k, using only source domain data for training and identical training procedures and configurations. This approach allows us to establish a suitable baseline and determine appropriate training parameters. First, we load the pretrained model and replace the linear classification head with one that matches the number of classes in our dataset (12 outputs for VisDa-2017 \cite{visda2017}, 102 outputs for Topex-Printer).
We perform three different training schemes: training the classification head only (CH), fine-tuning the full model (FT), and a combination of CH and FT, tuning the classification head first and continuing with full fine-tuning (CH-FT) inspired by \cite{lpft2022}. 
\begin{enumerate}
    \item For CH, we freeze all layers but the classification head and train for 20 epochs using SGD with learning rates [10.0, 1e-01, 1e-03], momentum 0.9, no weight decay, no learning rate scheduler, and no warmup.
    \item For FT, we do not freeze any layers and train for 20 epochs using AdamW optimizer with learning rates [1e-01, 1e-03, 1e-05], weight decay $0.01$, cosine annealing learning rate scheduler~\cite{sgdr2016} without restarts, and two warmup epochs (10\% of total epochs).
    \item For CH-FT, we use the best-performing CH training run based on the test set's top-1 accuracy and continue fine-tuning the whole model from the best validation checkpoint using parameters of the best-performing FT run for another 20 epochs (so 40 epochs total training after pretraining).
\end{enumerate}
For both datasets, VisDa-2017 and Topex-Printer, we use a batch size of 32 and two different data augmentation setups. For all runs, we use random resized crops with relative scale range (0.7, 1.0), random horizontal flip, random color jitter with parameters (brightness=0.3, contrast=0.3, saturation=0.3, hue=0.3), random grayscale, and normalize the final tensor using standard deviation [0.5, 0.5, 0.5] and mean [0.5, 0.5, 0.5]. We further replace random color jitter and random grayscale by AugMix~\cite{augmix2019} with default parameters. 

\subsection{Adapting to real-world images with unsupervised adaptation}
\label{sec:uda_experiments}
Upon completion of the first stage, we proceed with further experiments in an UDA setting. For these, we solely employ the SwinV2~\cite{swinv22021} and ViT~\cite{vit2020} model architectures, as these demonstrated superior performance (see supplementary material table \appref{apptab:evaluation-VisDA-2017} for details).
We start with the optimal classification head (CH) checkpoint from our experiments described in section~\ref{sec:dg_experiments} and keep the training parameters consistent with the best-performing fine-tuning (FT) run for each model. 
We execute six UDA runs each for the ViT~\cite{vit2020} and SwinV2~\cite{swinv22021} models: 20 training epochs, 32 batch size, AdamW optimizer with a 1e-05 learning rate, 1e-02 weight decay, and a cosine annealing learning rate scheduler without restarts and a two-epoch warmup (details in supplementary material). Image augmentations - random resized crop, horizontal flip, and AugMix~\cite{augmix2019} - are utilized as described in Sect.~\ref{sec:dg_experiments}.
Essentially, we replicate the process executed for source-domain-only CH-FT training runs, while concurrently incorporating UDA techniques — namely, CDAN~\cite{cdan2017} and MCC~\cite{mcc2019}.
Following the findings of~\cite{pretrainstudy2022} that CDAN~\cite{cdan2017} outperforms even newer DA techniques using modern architectures (Vit-L, ConvNext-XL), we decide to use the Transfer Learning Library (tllib)~\cite{transferability2022,tllib} implementations of CDAN (hidden size 1024) and MCC~\cite{mcc2019} (temperature 1.0) DA methods and also combine both.

\section{Evaluation}
\label{sec:eval}

\begin{figure}[tb]
    \includegraphics[width=0.495\linewidth]{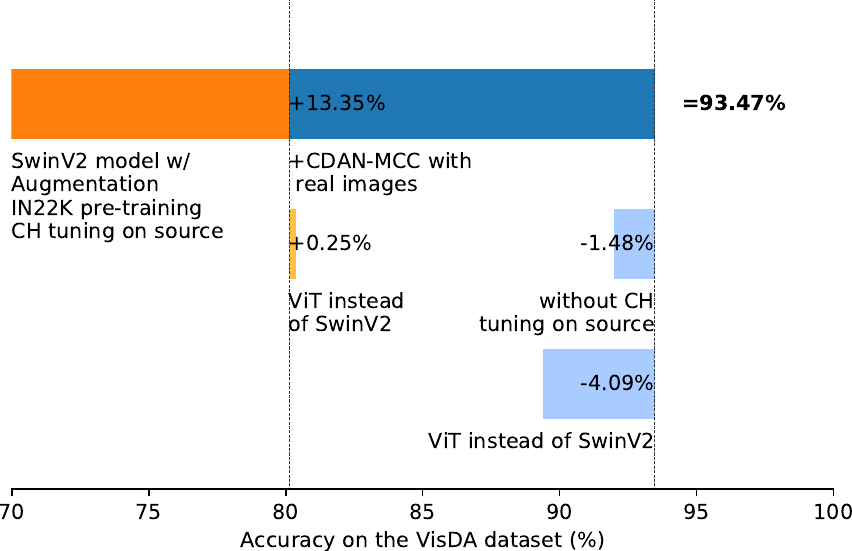}
    \includegraphics[width=0.495\linewidth]{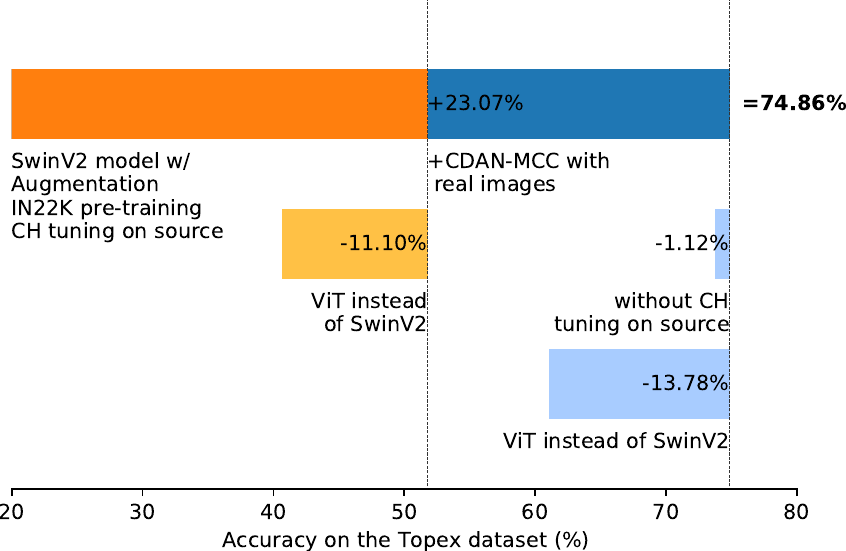}
    \centering
    \caption{
    Results of our DA pipeline for the (Left): VisDA and our (Right) Topex dataset. Blue bars highlight results
    obtained using UDA with additional target images.}
    \label{fig:performance-eval}
\end{figure}

\begin{table}[tb]
    \caption{Image classification top-1 accuracy in \% on VisDA-2017 target domain (real images) across all classes compared to literature. We report our best source-domain-only and UDA runs for the ViT and SwinV2 architecture. }
    \label{tab:compare-VisDa-2017}
    \begin{minipage}{\columnwidth}
        \centering
        \resizebox{\textwidth}{!}{
        \begin{tabular}{l|c|llllllllllll|l}
            \toprule
            Method &   & Pl & Bcl & Bus & Car & Hrs & Knf & Mcy & Per & Plt & Skb & Trn & Tck & Mean \\
            \midrule
            CDAN~\cite{cdan2017} & \parbox[t]{2mm}{\multirow{4}{*}{\rotatebox[origin=c]{90}{ResNet}}}
            & 85.2 & 66.9 & 83.0 & 50.8 & 84.2 & 74.9 & 88.1 & 74.5 & 83.4 & 76.0 & 81.9 & 38.0 & 73.9 \\
            MCC~\cite{mcc2019} &  & 88.1 & 80.3 & 80.5 & 71.5 & 90.1 & 93.2 & 85.0 & 71.6 & 89.4 & 73.8 & 85.0 & 36.9 & 78.8 \\
            SDAT~\cite{sdat2022} &  & 95.8 & 85.5 & 76.9 & 69.0 & 93.5 & 97.4 & 88.5 & 78.2 & 93.1 & 91.6 & 86.3 & 55.3 & 84.3 \\
            MIC~\cite{mic2022} &  & 96.7 & 88.5 & 84.2 & 74.3 & 96.0 & 96.3 & 90.2 & 81.2 & 94.3 & 95.4 & 88.9 & 56.6 & 86.9 \\
            \hline 
            TVT~\cite{tvt2021} & \parbox[t]{2mm}{\multirow{5}{*}{\rotatebox[origin=c]{90}{ViT}}} & 92.9 & 85.6 & 77.5 & 60.5 & 93.6 & 98.2 & 89.3 & 76.4 & 93.6 & 92.0 & 91.7 & 55.7 & 83.9 \\
            CDTRANS~\cite{cdtrans2021} &  & 97.1 & 90.5 & 82.4 & 77.5 & 96.6 & 96.1 & 93.6 & 88.6 & 97.9 & 86.9 & 90.3 & 62.8 & 88.4 \\
            SDAT~\cite{sdat2022} &  & 98.4 & 90.9 & 85.4 & 82.1 & 98.5 & 97.6 & 96.3 & 86.1 & 96.2 & 96.7 & 92.9 & 56.8 & 89.8 \\
            MIC~\cite{mic2022} &  & \textbf{99.0} & 93.3 & 86.5 & 87.6 & \textbf{98.9} & \textbf{99.0} & \textbf{97.2} & \textbf{89.8} & 98.9 & \textbf{98.9} & 96.5 & 68.0 & 92.8 \\
            \cline{1-1}
            Ours w/o UDA &  & 96.48 & 71.82 & 90.14 & \textbf{99.20} & 94.66 & 77.71 & 87.28 & 44.45 & 95.12 & 83.64 & 94.05 & 40.76 & 80.54 \\
            Ours &  & 94.82 & 93.49 & 92.80 & 95.89 & 90.95 & 88.51 & 77.46 & 75.42 & 96.27 & 97.32 & 94.74 & 88.03 & 89.38 \\
            \hline 
            Ours w/o UDA & \parbox[t]{2mm}{\multirow{2}{*}{\rotatebox[origin=c]{90}{Swin}}} & 97.09 & 80.48 & 85.35 & 98.12 & 92.39 & 83.54 & 94.85 & 19.89 & 89.13 & 78.89 & \textbf{97.03} & 55.18 & 80.12 \\
            Ours &  & 97.96 & \textbf{95.15} & \textbf{95.81} & 98.64 & 98.34 & 95.68 & 80.12 & 83.87 & \textbf{99.39} & 94.68 & 96.61 & \textbf{93.85} & \textbf{93.47} \\
            \bottomrule
        \end{tabular}
        }
        \bigskip\centering
        \footnotesize\
    \end{minipage}
\end{table}

Our experiments are always based on measuring the mean class-wise accuracy in the target domain, \textit{i.e.} the real-world images.

\myparagraph{Results on VisDA-2017 Dataset}
Our first evaluation is done on the standard domain adaptation benchmark VisDA-2017~\cite{visda2017}, where we are able to achieve 
SoTA performance as highlighted in Tab.~\ref{tab:compare-VisDa-2017}. One can see, that our ViT training outperforms TVT~\cite{tvt2021} and achieves competitive results compared to CDTRANS~\cite{cdtrans2021} and SDAT~\cite{sdat2022} but does not reach the performance of MIC~\cite{mic2022} when the same ViT architecture is used. %
However, our pipeline with the SwinV2 architecture slightly outperforms the current state of the art by 0.68\% accuracy.

Most importantly for us and the paper, we analyzed the contribution of each part of our pipeline in \figurename~\ref{fig:performance-eval} (left).
In this figure, the results of several ablations have been visualized with blueish bars referring to results achieved with additional target images through UDA techniques. The results reveal several
aspects:
\begin{enumerate}
\item Unsupervised domain adaptation is important to adapt to real-world images: Our best models with source data only, achieve around 80\% accuracy, but with CDAN~\cite{cdan2017} and MCC~\cite{mcc2019} as combined UDA techniques, we are
able to outperform all other approaches on this dataset.
\item It is beneficial and fast and easy to use class head (CH) tuning on the source data before applying UDA techniques to prevent feature distortion~\cite{lpft2022}: This can be seen in the $-1.48\%$ drop in performance without CH tuning. 
\item Using the right model architecture is crucial for UDA: Our ViT models after UDA achieve less than $90\%$ accuracy (drop of $4.09\%$). This difference in performance is insignificant before UDA.
\item Our SoTA performance was achieved after only 3 training epochs of fine-tuning from the pretrained checkpoint on a single Nvidia Tesla V100 PCIE 32GB GPU (CH-checkpoint after 1 epoch + 2 Epochs UDA with CDAN+MCC). However, the number of training epochs and training stability varies between our runs but almost all experiments achieve the best validation accuracy after just a few epochs of training.
\end{enumerate}

Further experimental results are given in the supplementary material of the paper and reveal the following additional aspects:
\begin{enumerate}
\item CDAN+MCC in combination outperforms CDAN and MCC individually in most cases (see supp. table \appref{apptab:evaluation-VisDA-2017-UDA} and table \appref{apptab:evaluation-topex-uda}).
\item Given the ConvNextV2~\cite{convnextv22023}-based runs' modest performance—12.42\% and 19.82\% for source-data-only experiments, we suspend further experiments with this architecture. (see supp. table \appref{apptab:eval_visda_all})
\end{enumerate}

\myparagraph{Results on the Topex-Printer Dataset}

The high accuracies on VisDA-2017~\cite{visda2017} in general and the marginal improvements achieved on this dataset in the last years, suggest the use of a more challenging dataset to benchmark domain
adaptation pipelines. Therefore, we developed and assembled the Topex-Printer dataset (Sect.~\ref{sec:topex_printer_dataset}).
The results on the dataset are given in \figurename~\ref{fig:performance-eval} (Right) and similar conclusions compared to the previous section can be drawn:
\begin{enumerate}
\item Unsupervised domain adaptation is even more important on this dataset: with a $23.07\%$ gain in performance, the domain gap between the rendered images and the real-world images is likely larger compared to VisDA-2017.
\item It is again reasonable to do CH tuning before UDA. Surprisingly, SwinV2 setups using CDAN~\cite{cdan2017} or MCC~\cite{mcc2019} alone do not benefit from using a tuned classification head but instead perform worse than just using the pretrained checkpoint from Huggingface (see supplementary material table~\appref{apptab:evaluation-topex-uda} for these results). However, when using CDAN and MCC combined starting from the tuned classification head, the final model performs 1.12\% better. For the ViT runs on the other hand, the CH initialized runs outperform runs without classification head tuning significantly.   
\item The Swin-V2 model shows a remarkable performance compared to the ViT model with a performance gain of $+11.10\%$ before UDA and $+13.78\%$ after UDA.
\end{enumerate}

\section{Conclusion}
We propose a practical approach for an image classifier in a DA setting using rendered images from 3D objects as the source domain and real images as the target domain. We conducted several experiments performing transfer learning with source data only to set a strong baseline for follow-up UDA training using the VisDA-2017 image classification challenge dataset and our newly proposed Topex-Printer dataset with more than 100 categories.
In our DA experiments, we outperformed the current state-of-the-art~\cite{mic2022} by achieving a mean accuracy of 93.47\% on the VisDA-2017 dataset and 74.86\% on the  Topex-Printer dataset. One goal in future work is to adapt our framework to object detection scenarios~\cite{goehring2014interactive}.

\myparagraph{Acknowledgements:} This work was funded by the German Federal Ministry of Education and Research (BMBF) through their support of the project SynthNet, a part of the KMU-Innovativ initiative (project code: 01IS21002C), the KI-Werkstatt project at the University of Applied Sciences Berlin (part of the Forschung an Fachhochschulen program (project code: 13FH028KI1) as well as project TAHAI (funded by IFAF Berlin).

\bibliographystyle{splncs04}  
\bibliography{paper}

\begin{thebibliography}{10}
\providecommand{\url}[1]{\texttt{#1}}
\providecommand{\urlprefix}{URL }
\providecommand{\doi}[1]{https://doi.org/#1}

\bibitem{in2009}
Deng, J., Dong, W., Socher, R., Li, L.J., Li, K., Fei-Fei, L.: Imagenet: A
  large-scale hierarchical image database. CVPR pp. 248--255 (2009)

\bibitem{vit2020}
Dosovitskiy, A., Beyer, L., Kolesnikov, A., Weissenborn, D., Zhai, X.,
  Unterthiner, T., Dehghani, M., Minderer, M., Heigold, G., Gelly, S., et~al.:
  An image is worth 16x16 words: Transformers for image recognition at scale.
  In: ICLR (2021)

\bibitem{dann2015}
Ganin, Y., Ustinova, E., Ajakan, H., Germain, P., Larochelle, H., Laviolette,
  F., Marchand, M., Lempitsky, V.: Domain-adversarial training of neural
  networks. The journal of machine learning research  \textbf{17}(1),
  2096--2030 (2016)

\bibitem{goehring2014interactive}
Goehring, D., Hoffman, J., Rodner, E., Saenko, K., Darrell, T.: Interactive
  adaptation of real-time object detectors. In: 2014 IEEE international
  conference on robotics and automation (ICRA). pp. 1282--1289. IEEE (2014)

\bibitem{GAN2014}
Goodfellow, I.J., Pouget-Abadie, J., Mirza, M., Xu, B., Warde-Farley, D.,
  Ozair, S., Courville, A.C., Bengio, Y.: Generative adversarial nets. In:
  NeurIPS (2014)

\bibitem{augmix2019}
Hendrycks, D., Mu, N., Cubuk, E.D., Zoph, B., Gilmer, J., Lakshminarayanan, B.:
  Augmix: A simple data processing method to improve robustness and
  uncertainty. In: ICLR (2019)

\bibitem{cycada2017}
Hoffman, J., Tzeng, E., Park, T., Zhu, J.Y., Isola, P., Saenko, K., Efros,
  A.A., Darrell, T.: Cycada: Cycle-consistent adversarial domain adaptation.
  In: ICML (2017)

\bibitem{mic2022}
Hoyer, L., Dai, D., Wang, H., Van~Gool, L.: Mic: Masked image consistency for
  context-enhanced domain adaptation. In: CVPR. pp. 11721--11732 (2023)

\bibitem{tllib}
Jiang, J., Chen, B., Fu, B., Long, M.: Transfer-learning-library.
  \url{https://github.com/thuml/Transfer-Learning-Library} (2020)

\bibitem{transferability2022}
Jiang, J., Shu, Y., Wang, J., Long, M.: Transferability in deep learning: A
  survey. ArXiv  \textbf{abs/2201.05867} (2022)

\bibitem{mcc2019}
Jin, Y., Wang, X., Long, M., Wang, J.: Minimum class confusion for versatile
  domain adaptation. In: ECCV (2019)

\bibitem{can2019}
Kang, G., Jiang, L., Yang, Y., Hauptmann, A.: Contrastive adaptation network
  for unsupervised domain adaptation. CVPR pp. 4888--4897 (2019)

\bibitem{pretrainstudy2022}
Kim, D., Wang, K., Sclaroff, S., Saenko, K.: A broad study of pre-training for
  domain generalization and adaptation. In: ECCV (2022)

\bibitem{lpft2022}
Kumar, A., Raghunathan, A., Jones, R.M., Ma, T., Liang, P.: Fine-tuning can
  distort pretrained features and underperform out-of-distribution. In: ICLR
  (2022)

\bibitem{swd2019}
Lee, C.Y., Batra, T., Baig, M.H., Ulbricht, D.: Sliced wasserstein discrepancy
  for unsupervised domain adaptation. CVPR pp. 10277--10287 (2019)

\bibitem{mscoco2014}
Lin, T.Y., Maire, M., Belongie, S.J., Hays, J., Perona, P., Ramanan, D.,
  Doll{\'a}r, P., Zitnick, C.L.: Microsoft coco: Common objects in context. In:
  ECCV (2014)

\bibitem{swinv22021}
Liu, Z., Hu, H., Lin, Y., Yao, Z., Xie, Z., Wei, Y., Ning, J., Cao, Y., Zhang,
  Z., Dong, L., Wei, F., Guo, B.: Swin transformer v2: Scaling up capacity and
  resolution. CVPR pp. 11999--12009 (2021)

\bibitem{dan2015}
Long, M., Cao, Y., Wang, J., Jordan, M.: Learning transferable features with
  deep adaptation networks. In: ICML. pp. 97--105. PMLR (2015)

\bibitem{cdan2017}
Long, M., Cao, Z., Wang, J., Jordan, M.I.: Conditional adversarial domain
  adaptation. In: NeurIPS (2017)

\bibitem{jan2016}
Long, M., Zhu, H., Wang, J., Jordan, M.I.: Deep transfer learning with joint
  adaptation networks. In: ICML (2016)

\bibitem{sgdr2016}
Loshchilov, I., Hutter, F.: Sgdr: Stochastic gradient descent with warm
  restarts. arXiv: Learning  (2016)

\bibitem{visda2017}
Peng, X., Usman, B., Kaushik, N., Wang, D., Hoffman, J., Saenko, K.: Visda: A
  synthetic-to-real benchmark for visual domain adaptation. In: CVPR-W. pp.
  2021--2026 (2018)

\bibitem{sdat2022}
Rangwani, H., Aithal, S.K., Mishra, M., Jain, A., Babu, R.V.: A closer look at
  smoothness in domain adversarial training. In: ICML (2022)

\bibitem{ytbb2017}
Real, E., Shlens, J., Mazzocchi, S., Pan, X., Vanhoucke, V.:
  Youtube-boundingboxes: A large high-precision human-annotated data set for
  object detection in video. CVPR pp. 7464--7473 (2017)

\bibitem{deit2021}
Touvron, H., Cord, M., Douze, M., Massa, F., Sablayrolles, A., Jegou, H.:
  Training data-efficient image transformers and distillation through
  attention. In: Meila, M., Zhang, T. (eds.) Proceedings of the 38th
  International Conference on Machine Learning. vol.~139, pp. 10347--10357.
  PMLR (18--24 Jul 2021)

\bibitem{adda2017}
Tzeng, E., Hoffman, J., Saenko, K., Darrell, T.: Adversarial discriminative
  domain adaptation. CVPR pp. 2962--2971 (2017)

\bibitem{convnextv22023}
Woo, S., Debnath, S., Hu, R., Chen, X., Liu, Z., Kweon, I.S., Xie, S.: Convnext
  v2: Co-designing and scaling convnets with masked autoencoders. ArXiv
  \textbf{abs/2301.00808} (2023)

\bibitem{cdtrans2021}
Xu, T., Chen, W., Pichao, W., Wang, F., Li, H., Jin, R.: Cdtrans: Cross-domain
  transformer for unsupervised domain adaptation. In: ICLR (2021)

\bibitem{tvt2021}
Yang, J., Liu, J., Xu, N., Huang, J.: Tvt: Transferable vision transformer for
  unsupervised domain adaptation. In: WACV. pp. 520--530 (2021)

\bibitem{cyclegan2017}
Zhu, J.Y., Park, T., Isola, P., Efros, A.A.: Unpaired image-to-image
  translation using cycle-consistent adversarial networks. In: ICCV (Oct 2017)

\end{thebibliography}

\ifthenelse{\boolean{combinedversion}}{
\newpage
\appendix

\section{Implementation Details}
\label{appsec:experiments}

\subsection{Adapting pretrained models to rendered images - implementation details}
\label{appsec:dg_experiments}
We use pretrained models \emph{"google/vit-base-patch16-224-in21k"} (ViT)~\cite{vit2020}, \emph{"microsoft/swinv2-base-patch4-window12-192-22k"} (SwinV2)~\cite{swinv22021}, \emph{"facebook/convnextv2-base-22k-224"} (ConvNextV2)~\cite{convnextv22023}, and \emph{"facebook/deit-base-distilled-patch16-224"} (DeiT)~\cite{deit2021} from Huggingface\footnote{https://huggingface.co/models} for experiments using the VisDA-2017 dataset but only ViT and SwinV2 for our Topex-Printer dataset. ViT, SwinV2, and ConvNextV2 were pretrained on ImageNet22K, while DeiT has been pretrained on ImagNet1K. 
We perform three different training schemes, training the classification head only (CH), fine-tuning the full model (FT), and a combination of CH and FT, tuning the classification head first and continuing with full fine-tuning (CH-FT) inspired by \cite{lpft2022}. 
\begin{enumerate}
    \item For CH we use the Pytorch\footnote{https://pytorch.org/} SGD optimizer with learning rates [10.0, 0.1, 0.001], momentum 0.9, no weight decay, no learning rate scheduler, and no warmup. 
    \item For FT we use the Pytorch implementation of AdamW optimizer with learning rates [0.1, 0.001, 0.00001], weight decay 0.01, cosine annealing learning rate scheduler\footnote{https://huggingface.co/docs/transformers/main\_classes/optimizer\_schedules} \cite{sgdr2016} without restarts, and two warmup epochs (10\% of total epochs). 
\end{enumerate}

For both datasets for data augmentation Pytorch 2.0.0 implementation \footnote{https://pytorch.org/vision/main/generated/torchvision.transforms.AugMix.html} is used. 

\subsection{Adapting to real-world images with unsupervised domain adaptation - implementation details}
\label{appsec:uda_experiments}
For UDA experiments we start from the best source-domain-only trained CH checkpoint with respect to the model architecture and continue training using the same parameters as the best FT run for each model as described in the paper. We use Pytorch 2.0.0 implementations of image augmentations random resized crop, horizontal flip, and AugMix~\cite{augmix2019} with the same parameters described in the last paragraph of section~\ref{appsec:dg_experiments}. We use the Transfer Learning Library (tllib)~\cite{transferability2022,tllib} implementations of CDAN (hidden size 1024) and MCC~\cite{mcc2019} (temperature 1.0) domain adaptation methods and also combine both using two different initial checkpoints for each model architecture. One initial checkpoint from Huggingface, pretrained on ImageNet22K~\cite{in2009} (\emph{"google/vit-base-patch16-224-in21k"} (ViT) and \emph{"microsoft/swinv2-base-patch4-window12-192-22k"} (SwinV2)) and the best-performing checkpoint after training only the classification head from our source-domain-only experiments. Again, we use global random seed 42 for all experiments and training is performed on a single Nvidia Tesla V100 PCIE 32GB GPU. 

Different from other methods, we perform considerably better correctly identifying the \emph{truck} class but underperform on the \emph{motorcycle} and \emph{person} class instead. The confusion matrix shown in figure~\ref{appfig:confmat} shows, that our trained model often mixes up motorcycle samples with bicycles (7\%) and skateboards (10\%) while the person class is mixed up rather uniformly (3\%-4\%) with skateboards, plants, motorcycles, and horses.     

\clearpage
\section{Dataset samples}
\begin{figure}[h!]
    \includegraphics[width=0.8\linewidth]{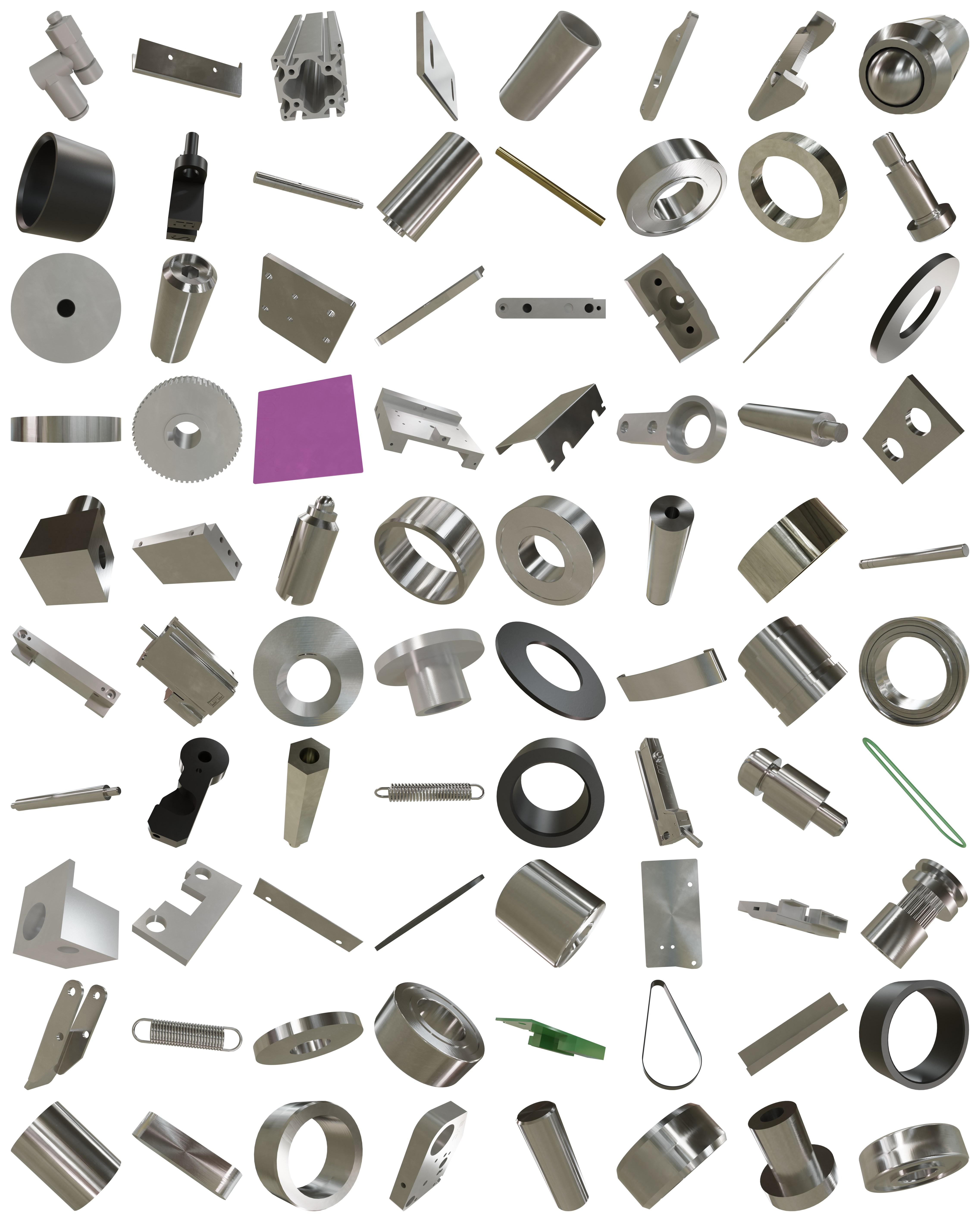}
    \centering
    \caption{
    80 random samples of rendered images from the Topex-Printer dataset. Each image $512^2$, featuring machine parts marked with bounding boxes, is trimmed according to these boxes, extended to form a rectangle, and padded with black if needed. Finally, all images are resized to a resolution of 256x256 pixels.
    }
    \label{appfig:topex_synth_samples}
\end{figure}

\begin{figure}[h!]
    \includegraphics[width=0.8\linewidth]{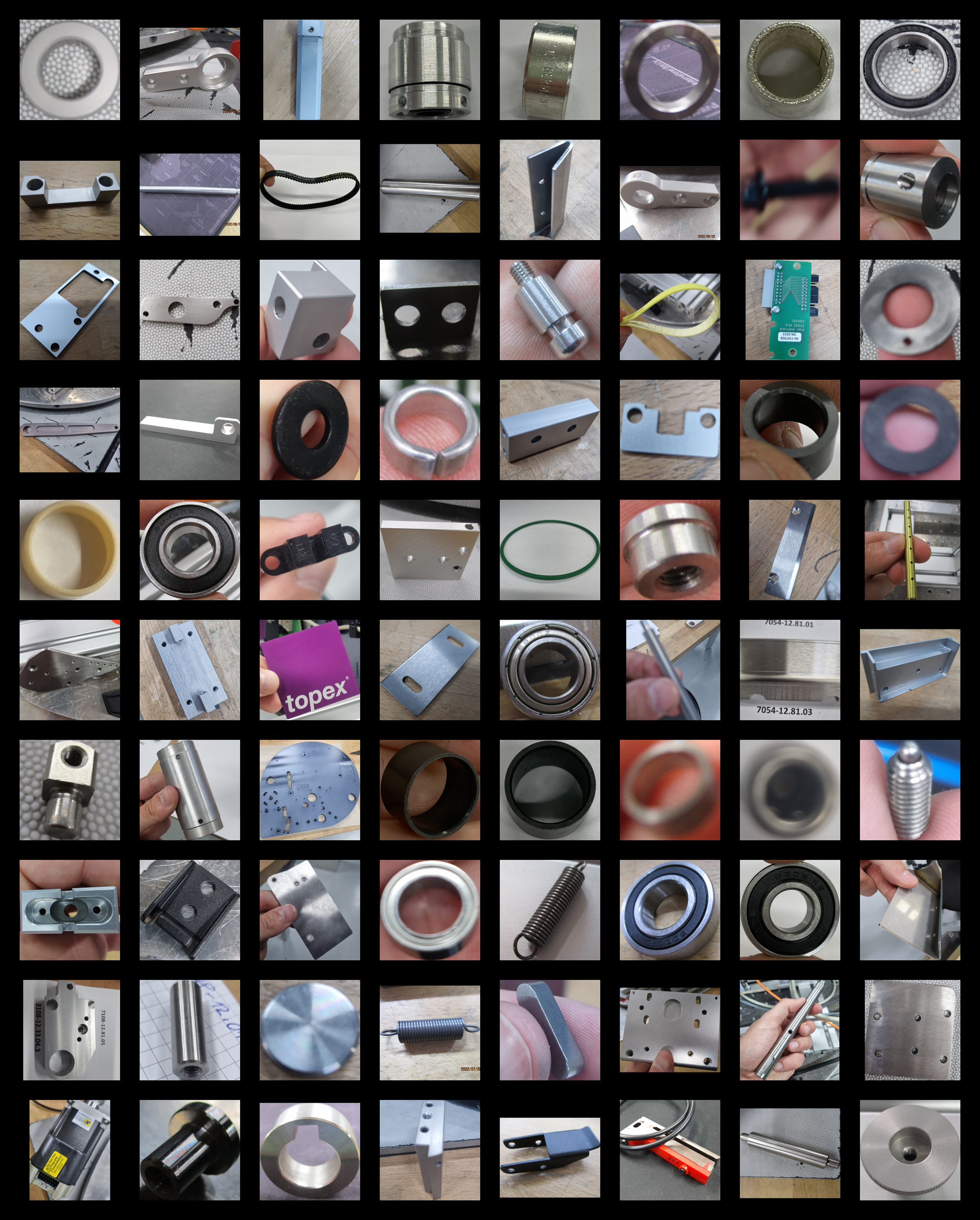}
    \centering
    \caption{
    80 random samples of real images from the Topex-Printer dataset.
    }
    \label{appfig:topex_real_samples}
\end{figure}

\begin{figure}[h!]
    \includegraphics[width=\linewidth]{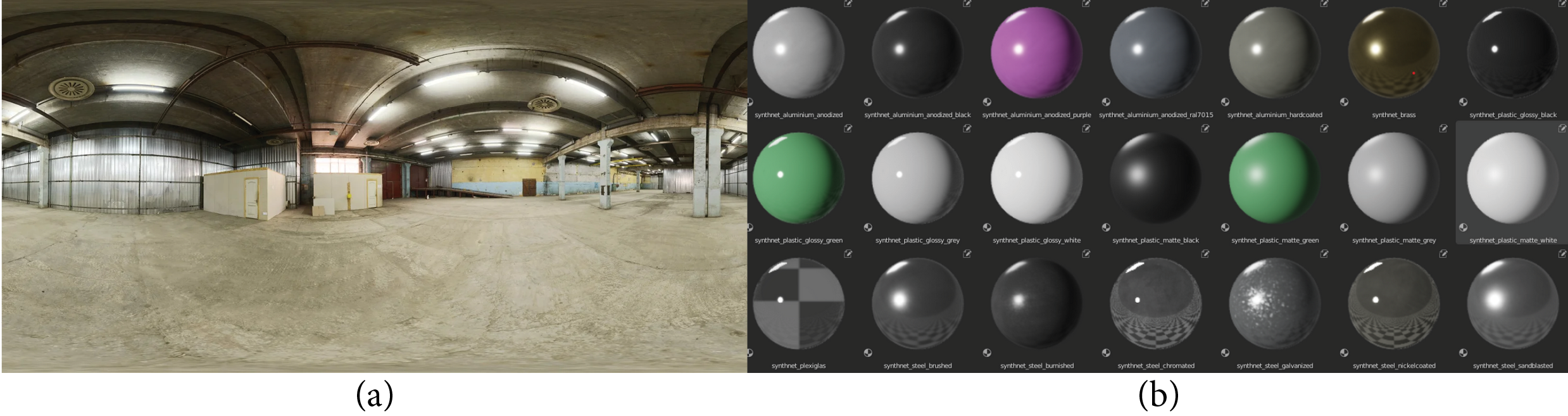}
    \centering
    \caption{
    (Best viewed in color) Left (a): HDRI of the warehouse environment map used in our rendering scene. Image by Sergej Majboroda [CC0], via Polyhaven.
    Right (b): Our handcrafted Blender material collection we used for the Topex-Printer dataset.
    }
    \label{appfig:envmap_materials}
\end{figure}

\clearpage
\section{Evaluation Results}
\begin{table}[h!]
    \caption{Acc@1 in \% on target domain (real images) for all source-domain-only training experiments on VisDA-2017 classification dataset. Note that \emph{base} transform means that random color jitter and random grayscale transforms are applied. Faded out rows are representing numerically instable runs that have been canceled due to NaN loss for example.}
    \label{apptab:eval_visda_all}
    \begin{minipage}{\columnwidth}
        \centering
        \begin{tabular}{lllll|r}
            \toprule
            \textbf{Model} & \textbf{Pre-training} & \textbf{train scheme} & \textbf{transform} & \textbf{lr} & \textbf{Acc@1} \\
            \midrule
            ViT-b & IN22K & CH & base & 1e+1 & 63.31 \\
            ViT-b & IN22K & CH & base & 1e-1 & 71.03 \\
            ViT-b & IN22K & CH & base & 1e-3 & 80.37 \\
            ViT-b & IN22K & CH & AugMix & 1e-3 & 80.18 \\
            \textcolor{gray}{ViT-b} & \textcolor{gray}{IN22K} & \textcolor{gray}{FT} & \textcolor{gray}{base} & \textcolor{gray}{1e-1} & \textcolor{gray}{07.64} \\
            ViT-b & IN22K & FT & base & 1e-3 & 17.69 \\
            ViT-b & IN22K & FT & base & 1e-5 & 66.88 \\
            ViT-b & IN22K & FT & AugMix & 1e-5 & 73.76 \\
            \textbf{ViT-b} & \textbf{IN22K} & \textbf{CH-FT} & \textbf{AugMix} & \textbf{1e-5} & \textbf{80.53} \\
            \hdashline
            SwinV2 & IN22K & CH & base & 1e+1 & 69.49 \\
            SwinV2 & IN22K & CH & base & 1e-1 & 72.02 \\
            SwinV2 & IN22K & CH & base & 1e-3 & 80.12 \\
            SwinV2 & IN22K & CH & AugMix & 1e-3 & 79.54 \\
            SwinV2 & IN22K & FT & base & 1e-3 & 18.84 \\
            SwinV2 & IN22K & FT & base & 1e-5 & 72.41 \\
            SwinV2 & IN22K & FT & AugMix & 1e-5 & 73.49 \\
            SwinV2 & IN22K & CH-FT & AugMix & 1e-5 & 76.96 \\
            \hdashline
            ConvNextV2 & IN22K & CH & base & 1e+1 & 12.81 \\
            ConvNextV2 & IN22K & CH & base & 1e-1 & 12.42 \\
            ConvNextV2 & IN22K & CH & base & 1e-3 & 11.30 \\
            ConvNextV2 & IN22K & CH & AugMix & 1e-3 & 11.98 \\
            \textcolor{gray}{ConvNextV2} & \textcolor{gray}{IN22K} & \textcolor{gray}{FT} & \textcolor{gray}{base} & \textcolor{gray}{1e-1} & \textcolor{gray}{10.04} \\
            ConvNextV2 & IN22K & FT & base & 1e-3 & 17.22 \\
            ConvNextV2 & IN22K & FT & base & 1e-5 & 19.82 \\
            ConvNextV2 & IN22K & FT & AugMix & 1e-5 & 11.98 \\
            ConvNextV2 CH-base-1e-3-e20 & IN22K & CH-FT & AugMix & 1e-5 & 25.43 \\
            \hdashline
            DeiT & IN1K & CH & base & 1e+1 & 59.21 \\
            DeiT & IN1K & CH & base & 1e-1 & 59.50 \\
            DeiT & IN1K & CH & base & 1e-3 & 75.13 \\
            \textcolor{gray}{DeiT} & \textcolor{gray}{IN1K} & \textcolor{gray}{FT} & \textcolor{gray}{base} & \textcolor{gray}{1e-1} & \textcolor{gray}{12.32} \\
            DeiT & IN1K & FT & base & 1e-3 & 21.00 \\
            DeiT & IN1K & FT & base & 1e-5 & 69.34 \\
            DeiT & IN1K & FT & AugMix & 1e-5 & 70.52 \\
            DeiT CH-base-1e-3-e20 & IN1K & CH-FT & AugMix & 1e-5 & 69.41 \\     
            DeiT CH-base-1e-3-e1 & IN1K & CH-FT & AugMix & 1e-5 & 74.12 \\
            \bottomrule
        \end{tabular}
        \bigskip\centering
        \footnotesize\
    \end{minipage}

\end{table}

\begin{table}[h!]
    \caption{Acc@1 in \% on target domain (real images) for all source-domain-only training experiments on the Topex-Printer dataset. Note that \emph{base} transform means that random color jitter and random grayscale transforms are applied. Faded out rows are representing numerically instable runs that have been canceled due to NaN loss for example.}
    \label{apptab:eval_topex_all}
    \begin{minipage}{\columnwidth}
        \centering
        \begin{tabular}{lllll|r}
            \toprule
            \textbf{Model} & \textbf{Pre-training} & \textbf{train scheme} & \textbf{transform} & \textbf{lr} & \textbf{Acc@1} \\
            \midrule
            ViT-b & IN22K & CH & base & 1e+1 & 34.85 \\
            ViT-b & IN22K & CH & base & 1e-1 & 40.69 \\
            ViT-b & IN22K & CH & base & 1e-3 & 31.78 \\
            \textcolor{gray}{ViT-b} & \textcolor{gray}{IN22K} & \textcolor{gray}{FT} & \textcolor{gray}{AugMix} & \textcolor{gray}{1e-1} & \textcolor{gray}{01.74} \\
            ViT-b & IN22K & FT & AugMix & 1e-3 & 21.75 \\
            ViT-b & IN22K & FT & AugMix & 1e-5 & 32.54 \\
            ViT-b & IN22K & CH-FT & AugMix & 1e-5 & 45.90 \\
            \hdashline
            SwinV2 & IN22K & CH & base & 1e+1 & 42.34 \\
            SwinV2 & IN22K & CH & base & 1e-1 & 45.15 \\
            SwinV2 & IN22K & CH & base & 1e-3 & 51.79 \\
            \textcolor{gray}{SwinV2} & \textcolor{gray}{IN22K} & \textcolor{gray}{FT} & \textcolor{gray}{AugMix} & \textcolor{gray}{1e-1} & \textcolor{gray}{01.70} \\
            SwinV2 & IN22K & FT & AugMix & 1e-3 & 26.23 \\
            SwinV2 & IN22K & FT & AugMix & 1e-5 & 25.69 \\
            SwinV2 & IN22K & CH-FT & AugMix & 1e-3 & 51.79 \\
            \textbf{SwinV2} & \textbf{IN22K} & \textbf{CH-FT} & \textbf{AugMix} & \textbf{1e-5} & \textbf{59.21} \\
            \bottomrule
        \end{tabular}
        \bigskip\centering
        \footnotesize\
    \end{minipage}
\end{table}

\begin{table}[h!]
    \caption{Acc@1 in \% on target domain (real images) for best results per model and training scheme in our source domain training experiments on VisDA-2017 classification dataset. Note that \emph{base} transform means that random color jitter and random grayscale transforms are applied instead of AugMix (other augmentations stay the same as explained in section~\ref{appsec:dg_experiments}).
    }
    \label{apptab:evaluation-VisDA-2017}
    \begin{minipage}{\columnwidth}
        \centering
        \begin{tabular}{lllll|r}
            \toprule
            \textbf{Model} & \textbf{Pre-training} & \textbf{train scheme} & \textbf{transform} & \textbf{lr} & \textbf{Acc@1} \\
            \midrule
            ViT-b & IN22K & CH & base & 1e-3 & 80.37 \\
            ViT-b & IN22K & FT & AugMix & 1e-5 & 73.76 \\
            \textbf{ViT-b} & \textbf{IN22K} & \textbf{CH-FT} & \textbf{AugMix} & \textbf{1e-5} & \textbf{80.53} \\
            \hdashline
            SwinV2 & IN22K & CH & base & 1e-3 & 80.12 \\
            SwinV2 & IN22K & FT & AugMix & 1e-5 & 73.49 \\
            SwinV2 & IN22K & CH-FT & AugMix & 1e-5 & 76.96 \\
            \hdashline
            DeiT & IN1K & CH & base & 1e-3 & 75.13 \\
            DeiT & IN1K & FT & AugMix & 1e-5 & 70.52 \\
            DeiT & IN1K & CH-FT & AugMix & 1e-5 & 74.12 \\
            \bottomrule
        \end{tabular}
        \bigskip\centering
        \footnotesize\
    \end{minipage}
\end{table}

\begin{table}[h!]
    \caption{Acc@1 in \% on target domain (real images) for best results per model and training scheme in our source-domain-only training experiments on Topex-Printer dataset. Note that \emph{base} transform means that random color jitter and random grayscale transforms are applied instead of AugMix (other augmentations stay the same as explained in section~\ref{appsec:dg_experiments}).
    }
    \label{apptab:evaluation-topex}
    \begin{minipage}{\columnwidth}
        \centering
        \begin{tabular}{lllll|r}
            \toprule
            \textbf{Model} & \textbf{Pre-training} & \textbf{train scheme} & \textbf{transform} & \textbf{lr} & \textbf{Acc@1} \\
            \midrule
            ViT-b & IN22K & CH & base & 1e-1 & 40.69 \\
            ViT-b & IN22K & FT & AugMix & 1e-5 & 32.54 \\
            ViT-b & IN22K & CH-FT & AugMix & 1e-5 & 45.90 \\
            \hdashline
            SwinV2 & IN22K & CH & base & 1e-3 & 51.79 \\
            SwinV2 & IN22K & FT & AugMix & 1e-5 & 25.69 \\
            \textbf{SwinV2} & \textbf{IN22K} & \textbf{CH-FT} & \textbf{AugMix} & \textbf{1e-5} & \textbf{59.21} \\
            \bottomrule
        \end{tabular}
        \bigskip\centering
        \footnotesize\
    \end{minipage}
\end{table}

\begin{table}[h!]
    \caption{Acc@1 in \% on target domain (real images) for all UDA experiments on VisDA-2017 classification dataset. Note that \emph{init checkpoint} describes the model checkpoint used for the UDA experiments. CH refers to the best-performing CH training scheme from our DG experiments respecting the used model architecture and IN22K refers to the respective Huggingface model checkpoints described in section~\ref{appsec:uda_experiments}.}
    \label{apptab:evaluation-VisDA-2017-UDA}
    \begin{minipage}{\columnwidth}
        \centering
        \begin{tabular}{lll|r}
            \toprule
            \textbf{Model} & \textbf{DA method} & \textbf{init checkpoint} & \textbf{Acc@1} \\
            \midrule
            ViT-b & CDAN & IN22K & 61.96 \\
            ViT-b & CDAN & CH & 88.78 \\
            ViT-b & MCC & IN22K & 79.63 \\
            ViT-b & MCC & CH & 88.88 \\
            ViT-b & CDAN-MCC & IN22K & 75.26 \\
            ViT-b & CDAN-MCC & CH & 89.38 \\
            \hdashline
            SwinV2 & CDAN & IN22K & 71.21 \\
            SwinV2 & CDAN & CH & 80.12 \\
            SwinV2 & MCC & IN22K & 90.65 \\
            SwinV2 & MCC & CH & 91.88 \\
            SwinV2 & CDAN-MCC & IN22K & 91.99 \\
            \textbf{SwinV2} & \textbf{CDAN-MCC} & \textbf{CH} & \textbf{93.47} \\
            \bottomrule
        \end{tabular}
        \bigskip\centering
        \footnotesize\
    \end{minipage}
\end{table}

\begin{table}[h!]
    \caption{Acc@1 in \% on target domain (real images) for all UDA experiments on the Topex-Printer dataset. Note that \emph{init checkpoint} describes the model checkpoint used for the UDA experiments. \emph{CH} refers to the best-performing CH training scheme from our source-domain-only training experiments respecting the used model architecture and IN22K refers to the respective Huggingface model checkpoints described in section~\ref{appsec:uda_experiments}.}
    \label{apptab:evaluation-topex-uda}
    \begin{minipage}{\columnwidth}
        \centering
        \begin{tabular}{lll|r}
            \toprule
            \textbf{Model} & \textbf{DA method} & \textbf{init checkpoint} & \textbf{Acc@1} \\
            \midrule
            ViT-b & CDAN & IN22K & 43.31 \\
            ViT-b & CDAN & CH & 47.51 \\
            ViT-b & MCC & IN22K & 32.95 \\
            ViT-b & MCC & CH & 61.36 \\
            ViT-b & CDAN-MCC & IN22K & 43.33 \\
            ViT-b & CDAN-MCC & CH & 61.08 \\
            \hdashline
            SwinV2 & CDAN & IN22K & 65.51 \\
            SwinV2 & CDAN & CH & 61.94 \\
            SwinV2 & MCC & IN22K & 72.86 \\
            SwinV2 & MCC & CH & 71.14 \\
            SwinV2 & CDAN-MCC & IN22K & 73.74 \\
            \textbf{SwinV2} & \textbf{CDAN-MCC} & \textbf{CH} & \textbf{74.86} \\
            \bottomrule
        \end{tabular}
        \bigskip\centering
        \footnotesize\
    \end{minipage}
\end{table}

\begin{figure}[h!]
    \includegraphics[width=0.7\linewidth]{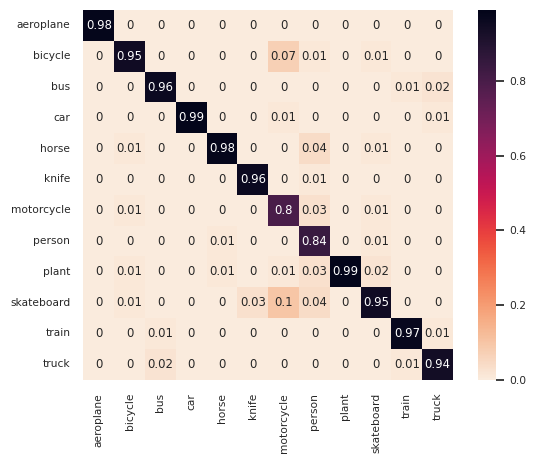}
    \centering
    \caption{
    Confusion matrix for our best-performing model on VisDA-2017: SwinV2-CH-CDAN-MCC
    }
    \label{appfig:confmat}
\end{figure}

\begin{table}[h!]
    \caption{Image classification top-1 accuracy in \% on VisDA-2017 target domain (real images) across all classes compared to literature. We report our best source-domain-only and UDA runs for the ViT and SwinV2 architecture. }
    \label{apptab:compare-VisDA-2017}
    \begin{minipage}{\columnwidth}
        \centering
        \resizebox{\textwidth}{!}{
        \begin{tabular}{l|c|llllllllllll|l}
            \toprule
            Method &   & Pl & Bcl & Bus & Car & Hrs & Knf & Mcy & Per & Plt & Skb & Trn & Tck & Mean \\
            \midrule
            CDAN~\cite{cdan2017} & \parbox[t]{2mm}{\multirow{4}{*}{\rotatebox[origin=c]{90}{ResNet}}}
            & 85.2 & 66.9 & 83.0 & 50.8 & 84.2 & 74.9 & 88.1 & 74.5 & 83.4 & 76.0 & 81.9 & 38.0 & 73.9 \\
            MCC~\cite{mcc2019} &  & 88.1 & 80.3 & 80.5 & 71.5 & 90.1 & 93.2 & 85.0 & 71.6 & 89.4 & 73.8 & 85.0 & 36.9 & 78.8 \\
            SDAT~\cite{sdat2022} &  & 95.8 & 85.5 & 76.9 & 69.0 & 93.5 & 97.4 & 88.5 & 78.2 & 93.1 & 91.6 & 86.3 & 55.3 & 84.3 \\
            MIC~\cite{mic2022} &  & 96.7 & 88.5 & 84.2 & 74.3 & 96.0 & 96.3 & 90.2 & 81.2 & 94.3 & 95.4 & 88.9 & 56.6 & 86.9 \\
            \hline 
            TVT~\cite{tvt2021} & \parbox[t]{2mm}{\multirow{5}{*}{\rotatebox[origin=c]{90}{ViT}}} & 92.9 & 85.6 & 77.5 & 60.5 & 93.6 & 98.2 & 89.3 & 76.4 & 93.6 & 92.0 & 91.7 & 55.7 & 83.9 \\
            CDTRANS~\cite{cdtrans2021} &  & 97.1 & 90.5 & 82.4 & 77.5 & 96.6 & 96.1 & 93.6 & 88.6 & 97.9 & 86.9 & 90.3 & 62.8 & 88.4 \\
            SDAT~\cite{sdat2022} &  & 98.4 & 90.9 & 85.4 & 82.1 & 98.5 & 97.6 & 96.3 & 86.1 & 96.2 & 96.7 & 92.9 & 56.8 & 89.8 \\
            MIC~\cite{mic2022} &  & \textbf{99.0} & 93.3 & 86.5 & 87.6 & \textbf{98.9} & \textbf{99.0} & \textbf{97.2} & \textbf{89.8} & 98.9 & \textbf{98.9} & 96.5 & 68.0 & 92.8 \\
            \cline{1-1}
            Source Only &  & 96.48 & 71.82 & 90.14 & \textbf{99.20} & 94.66 & 77.71 & 87.28 & 44.45 & 95.12 & 83.64 & 94.05 & 40.76 & 80.54 \\
            Ours &  & 94.82 & 93.49 & 92.80 & 95.89 & 90.95 & 88.51 & 77.46 & 75.42 & 96.27 & 97.32 & 94.74 & 88.03 & 89.38 \\
            \hline 
            Source Only & \parbox[t]{2mm}{\multirow{2}{*}{\rotatebox[origin=c]{90}{Swin}}} & 97.09 & 80.48 & 85.35 & 98.12 & 92.39 & 83.54 & 94.85 & 19.89 & 89.13 & 78.89 & \textbf{97.03} & 55.18 & 80.12 \\
            Ours &  & 97.96 & \textbf{95.15} & \textbf{95.81} & 98.64 & 98.34 & 95.68 & 80.12 & 83.87 & \textbf{99.39} & 94.68 & 96.61 & \textbf{93.85} & \textbf{93.47} \\
            \bottomrule
        \end{tabular}
        }
        \bigskip\centering
        \footnotesize\
    \end{minipage}
\end{table}

}{}

\end{document}